\documentclass[10pt,twocolumn,letterpaper]{article}

\usepackage{cvpr}              

\usepackage{graphicx}
\usepackage{amsmath}
\usepackage{amssymb}
\usepackage{booktabs}

\usepackage[accsupp]{axessibility}

\usepackage{times}
\usepackage{epsfig}
\usepackage{graphicx}
\usepackage{amsmath}
\usepackage{amssymb}

\usepackage{multirow}
\usepackage{booktabs}
\usepackage{subfloat}
\usepackage{xcolor}
\usepackage[switch]{lineno}
\usepackage{soul}

\usepackage{algpseudocode}
\usepackage{algorithm}

\usepackage[english]{babel}
\usepackage{amsthm}

\usepackage{caption}
\usepackage{subcaption}

\newtheorem{theorem}{Theorem}
\newtheorem{remark}{Remark}

\usepackage{xspace}
\makeatletter
\DeclareRobustCommand\onedot{\futurelet\@let@token\@onedot}
\def\@onedot{\ifx\@let@token.\else.\null\fi\xspace}

\makeatother

\def \py {\textcolor{red}}

\usepackage[pagebackref,breaklinks,colorlinks]{hyperref}

\usepackage[capitalize]{cleveref}
\crefname{section}{Sec.}{Secs.}
\Crefname{section}{Section}{Sections}
\Crefname{table}{Table}{Tables}
\crefname{table}{Tab.}{Tabs.}

\def\cvprPaperID{11756} 
\def\confName{CVPR}
\def\confYear{2022}

\begin{document}

\title{Discrete Wasserstein Distributional Matching for Quantization in Image Hashing}
\title{One Loss for Quantization: \\Deep Hashing with Discrete Wasserstein Distributional Matching}

\author{Khoa D. Doan, Peng Yang, Ping Li\\
Cognitive Computing Lab\\
Baidu Research\\
10900 NE 8th St. Bellevue, WA 98004, USA\\
{\tt\small \{khoadoan106, pengyang01,  pingli98\}@gmail.com}
}
\maketitle

\begin{abstract}\vspace{-0.12in}
Image hashing is a principled approximate nearest neighbor approach to find similar items to a query in a large collection of images. 
Hashing aims to learn a binary-output function that maps an image to a binary vector. 
For optimal retrieval performance, producing balanced hash codes with low-quantization error to bridge the gap between the learning stage's continuous relaxation and the inference stage's discrete quantization is important. 
However, in the existing deep supervised hashing methods, coding balance and low-quantization error are difficult to achieve and involve several losses.
We argue that this is because the existing quantization approaches in these methods are heuristically constructed and not effective to achieve these objectives. This paper considers an alternative approach to learning the quantization constraints. The task of learning balanced codes with low quantization error is re-formulated as matching the learned distribution of the continuous codes to a pre-defined discrete, uniform distribution. This is equivalent to minimizing the distance between two distributions. We then propose a computationally efficient distributional distance by leveraging the discrete property of the hash functions. This  distributional distance is a valid distance and enjoys lower time and sample complexities. The proposed single-loss quantization objective can be integrated into any existing supervised hashing method to improve code balance and quantization error. Experiments confirm that the proposed approach substantially improves the  performance of several representative hashing~methods. \vspace{-0.17in}
\end{abstract}

\section{Introduction}\label{sec:intro}
\vspace{-0.02in}

An important challenge associated with massive image datasets is to efficiently and effectively search for images containing \textit{semantically} \textbf{similar} content in these datasets. Hashing is a principled approximate nearest neighbor search approach with applications in many domains, ranging from text or image retrieval~\cite{salakhutdinov2009semantic,gong2013iterative} to spam or duplicate-scene detection~\cite{li2011hashing,chum2008near}. Hashing approaches learn binary encoding of the original images so that the ``candidate'' subset of images can be efficiently discovered from the binary-coding space. Binary codes are efficient to store and the high cost of pairwise distance calculations in the high-dimensional space is reduced to the significantly lower cost of discrete Hamming distance calculations. A Hamming-distance calculation only requires a bit-wise XOR and a bit-count operation, which can be efficiently computed in most \py{of} conventional systems.

To ensure that the retrieved images are relevant, hashing methods learn hash functions that preserve the pairwise similarity of the images in the discrete space. Supervised hashing methods additionally leverage the annotated similarity to learn the hash functions and achieve superior retrieval performance compared to unsupervised hashing methods~\cite{shen2015supervised,yang2018supervised,ge2014graph,xia2014supervised,cao2018hashgan,li2017deep,cao2018deep,su2018greedy,zheng2020deep}. Since discrete optimization is intractable, these methods solve a relaxed problem that replaces the discrete constraint with a continuous output. The continuous output is ``quantized'' to obtain the binary during inference. Such a relaxation results in a discrepancy between the discrete and continuous optimizations that must be compensated for in the learning process. 
Two important criteria to consider are quantization error and coding balance~\cite{weiss2009spectral,gong2013iterative,wang2017survey}. Quantization error is the information loss when the discrete function is represented by a continuous function. Quantization error penalizes the cases of assigning ``very'' similar data points to binary codes with large distances~\cite{weiss2009spectral}. Coding balance, on the other hand, encourages a uniform distribution of the images into the binary codes, which helps reduce the time complexity of the retrieval operations in the worst and average scenarios~\cite{he2011compact}.

Existing supervised hashing methods, especially  those that are based on neural networks, include one or more penalty terms, besides the similarity-preserving loss, to force the continuous output as discrete as possible. 
However, these relaxation schemes still introduce non-trivial quantization error and coding unbalanced, which eventually leads to sub-optimal hash codes~\cite{weiss2009spectral,wang2017survey}. Multiple penalty terms also lead to longer model trainings due to the time-consuming hyperparameter-tuning step. 

This paper proposes a faster and more performant quantization approach for the deep supervised hashing methods. First, we empirically show that low-quantization error and balance coding induce a uniform discrete distribution. The ultimate goal of hash-function learning is to project data into this uniform discrete distribution. Thus, we formulate the quantization objective as minimizing the distributional distance between the relaxed, continuous hash distribution and this uniform discrete distribution. 
The proposed formulation has two advantages: (i) achieving low-quantization error and coding balance is easier with this formulation and (ii) low-quantization error and coding balance are simultaneously optimized in a unified formulation, thus reducing the number of hyperparameters to tune. Our \textbf{main contributions} are as follows:

\begin{itemize}
    \item We achieve \py{a} low quantization error and coding balance by minimizing the single-loss distributional distance between the learned hash distribution and the uniform discrete distribution. This new quantization objective can be used in conjunction with any  existing deep supervised hashing methods to further improve their retrieval performances and reduce their time-consuming hyperparameter tuning processes.
    \item We propose a low computation- and sample-complexity Sliced-Wasserstein-based distributional distance. The proposed distance, called HSWD, is theoretically a valid distance with better computational efficiency than other Wasserstein-distance types, including the original Sliced Wasserstein Distance.
    \item We demonstrate the efficiency and effectiveness of the proposed quantization technique in several well-known deep supervised hashing methods on various widely used real-world datasets using both quantitative and qualitative performance analysis.
\end{itemize}

\noindent \textbf{Paper Organization:} The rest of the paper is organized as follows. We review the related work in Section~\ref{sec:related_work}. In Section~\ref{sec:methodology}, we present the empirical analysis of the quantization error and coding balance, and the details of the proposed methodology. We evaluate the effectiveness of the proposed quantization approach in Section~\ref{sec:experiments}. Finally, Section~\ref{sec:conclusions} presents remarks and concludes this paper. We present more details about experimental settings and results as well as supporting proofs in the supplementary material. 

\section{Related Work}\label{sec:related_work}

In this section, we first discuss the prior works in deep supervised hashing and the quantization constraints. Then we discuss the Wasserstein distances, which are related to the computational approach proposed in the paper.

\subsection{Image Hashing}

Hashing has been intensively investigated in both theory and practice, with applications ranging from retrieval~\cite{salakhutdinov2009semantic} to compressed sensing~\cite{Proc:Li_AISTATS17} and feature learning~\cite{Proc:Li_AISTATS21}. 
In the image retrieval domain, existing data-dependent hashing methods can be organized into two categories: shallow hashing and deep hashing. Shallow hashing methods rely on hand-crafted features and learn linear hash functions and feature extraction techniques~\cite{gong2013iterative,weiss2009spectral,kulis2009learning,liu2012supervised}. Along with the rapid development of deep neural networks, deep hashing methods~\cite{cao2017hashnet,lin2015deep,yang2019distillhash,yang2018semantic} combine representation learning and hash-function learning into an end-to-end model and have demonstrated significant performance improvements over the hand-crafted shallow hashing approaches.  Hashing methods can also be broadly classified into (data dependent) unsupervised~\cite{huang2017unsupervised,lin2016learning,huang2016unsupervised,do2016learning,he2013k,heo2012spherical,gong2013iterative,weiss2009spectral,salakhutdinov2009semantic,yang2019distillhash,doan2020imagehash,ghasedi2018unsupervised} and supervised methods~\cite{shen2015supervised,yang2018supervised,ge2014graph,xia2014supervised,cao2018hashgan,li2017deep,cao2018deep,yuan2020central,cao2017hashnet}. Supervised hashing methods demonstrate superior performance over the unsupervised ones by utilizing the labeled similarity information in the training data to learn the hash functions. Note that there is also a popular line of hashing works on ``data independent" (unsupervised) hashing with quantizations, including (i) quantized random projections~\cite{Article:Goemans_JACM95,Proc:Charikar_STOC02,Book:Vempala04,Proc:Datar_SCG04,Proc:Li_ICML14,Proc:Li_NIPS17}; (ii) quantized stable random projections~\cite{Proc:Li_AISTATS17}; (iii) quantized random Fourier features~\cite{Proc:Zhang_AISTATS19,Proc:Li_AISTATS21}; (iv) b-bit minwise hashing~\cite{Proc:Broder_WWW97,Proc:Li_Konig_WWW10} and b-bit consistent weighted sampling~\cite{Report:Manasse_CWS10,Proc:Ioffe_ICDM10,Proc:Li_KDD17}, etc.

Since learning the discrete hash function is computationally intractable, existing deep hashing methods approximate the discrete hash function with a continuous-output one, by replacing the discrete-output constraint with a $tanh$ activation~\cite{cao2017hashnet,yuan2020central,zheng2020deep,li2017deep}. To learn the hash functions, these methods optimize the similarity-preserving loss and quantization loss, simultaneously. The similarity-preserving loss tries to preserve the input-space similarity structure in the discrete space while the quantization loss minimizes the discrepancy between the discrete and continuous optimizations. It has been shown that an effective quantization approach is crucial in improving the retrieval performance~\cite{weiss2009spectral,doan2020efficient,Proc:Li_ICML14}. However, such quantization objective is often ignored or not effectively learned via multiple losses ($>3$) in the existing deep supervised hashing methods. This paper studies the effectiveness of quantization objectives in these approaches and proposes an alternative, more effective single-loss quantization objective. Note that, we consider our work to be orthogonal to OrthoHash~\cite{hoe2021one} although OrthoHash also aims to reduce the number of losses (but only in the point-wise supervised image hashing). Our approach does not require Batch Normalization for coding balance as in OrthoHash, can be used in the point-wise or pair-wise setting, and can be incorporated into any existing hashing methods to improve its quantization effectiveness.

\subsection{Wasserstein Distances}

\begin{figure*}[!htpb]
    \centering
    \includegraphics[width=0.95 \textwidth]{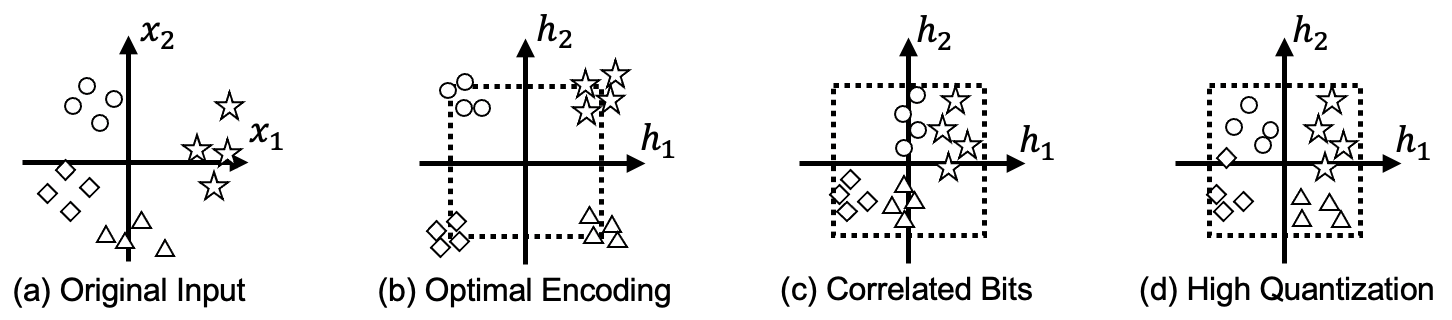}
    \vspace{-5pt}
    \caption{A visual illustration of the optimal function (Figure (b)) and learned hash function with poor the code balance (Figure (c)) and quantization constraint (Figure (d)). Figure (a) shows the original data clusters from four classes.}
    \label{fig:demo_toy}
\end{figure*}

Wasserstein distances have been extensively studied in the literature and especially in recent applications such as GANs~\cite{goodfellow2014generative,arjovsky2017wasserstein,doan2020image}. One advantage of the Wasserstein distances over other types of distributional distances, such as Kullback–Leibler (KL) or Jensen-Shannon (JS) divergence, is that they consider the geometry of the data and are well-defined even for distributions without overlapping supports~\cite{arjovsky2017towards}. However, estimating the Wasserstein distance is a challenging task. Computing the Wasserstein-2 distance from its primal domain is computationally intractable~\cite{arjovsky2017wasserstein}. Computing the Wasserstein-2 distance via the Optimal-Transport (OT) formulation has also been explored~\cite{doan2021interpretable,doan2020efficient}, but the OT's high computational cost is prohibitive. Employing the Kantorovich-Rubinstein dual requires estimating a Lipschitz continuous function to calculate the distance. Parameterizing the Lipschitz function with a neural network is challenging~\cite{deshpande2019max}. Furthermore, the dual approach results in a minimax optimization, which requires non-trivial modifications to learning algorithms such as those in hashing. Note that, estimating the KL or JS divergence also requires minimax optimization~\cite{doan2019adversarial}.

Wasserstein-2 distance (also KL and JS) requires an exponential number of samples to reliably estimate the distance~\cite{deshpande2018generative}. In contrast, a variant of the Wasserstein distance, called Sliced Wasserstein Distance (SWD), approximates the Wasserstein distance by averaging the one-dimensional Wasserstein distances of the data points when they are projected onto many random, one-dimensional directions ~\cite{kolouri2019generalized}. SWD has a polynomial sample complexity~\cite{deshpande2019max}. The SWD estimation has a computational cost of $O(LN \log(Nd))$, where $L$ is the number of random directions, $N$ is the number of samples, and $d$ is the dimension of the data.  Nevertheless, in the high dimensional space, it becomes very likely that many random, one-dimensional directions do not lie on the manifold of the data. In other words, in several of these directions, the projected distances are close to zero. Consequently, in practice, the number of random directions $L$ is often large. For example, in~\cite{deshpande2018generative}, for a mini-batch size of $64$, SWD needs $L=10,000$ projections. To address this problem, some works find the best directions and estimate the Wasserstein distance using these directions~\cite{deshpande2019max,doan2021backdoor,doan2020imagehash}.

\section{Distributional Matching for Hash Function Learning}\label{sec:methodology}

\subsection{Hash Function Learning}

\begin{figure*}[!htpb]
	\centering
	\begin{subfigure}[t]{0.23\textwidth}
		\centering
		\includegraphics[width=1.0 \textwidth]{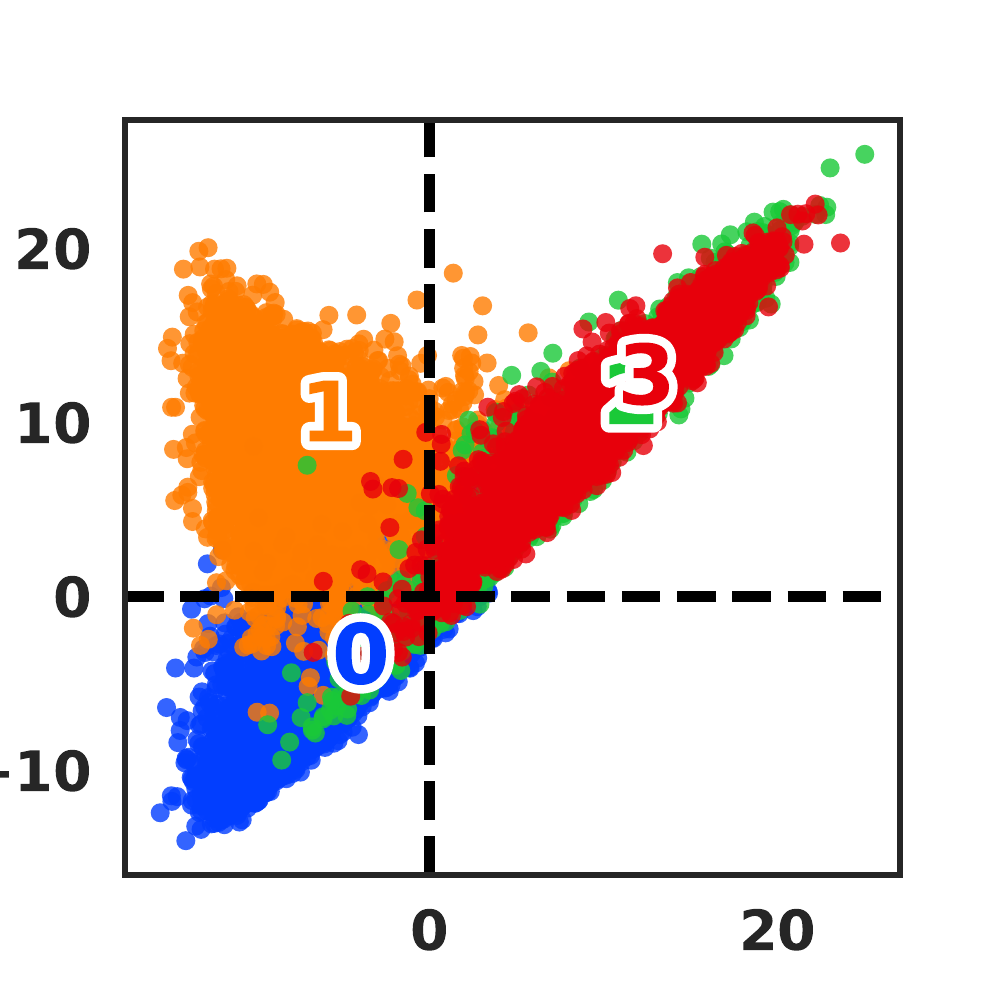}
		\caption{HashNet (mAP: 0.7208)}\label{fig:1a}
	\end{subfigure}
	\quad
	\begin{subfigure}[t]{0.23\textwidth}
		\centering
		\includegraphics[width=1.0 \textwidth]{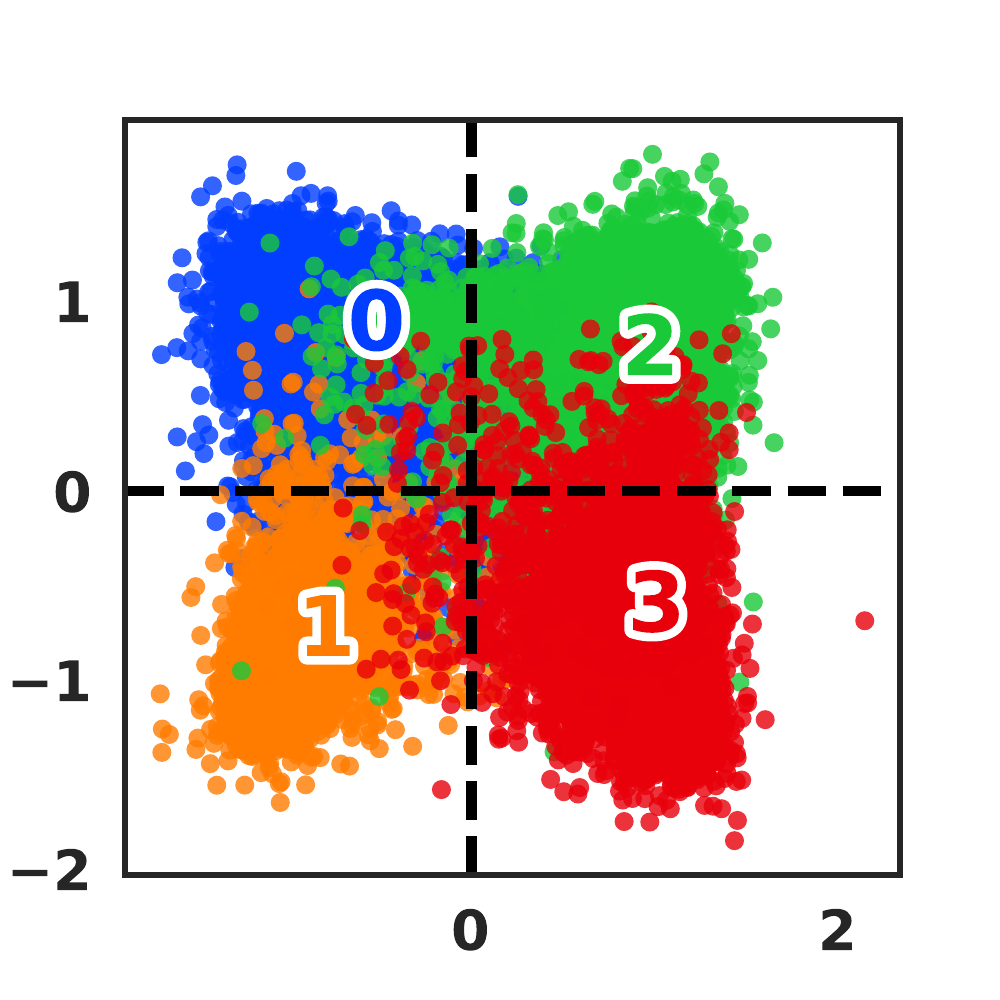}
		\caption{HashNet-C (mAP: 0.8500)}\label{fig:1b}
	\end{subfigure}
	\quad
	\begin{subfigure}[t]{0.23\textwidth}
		\centering
		\includegraphics[width=1.0 \textwidth]{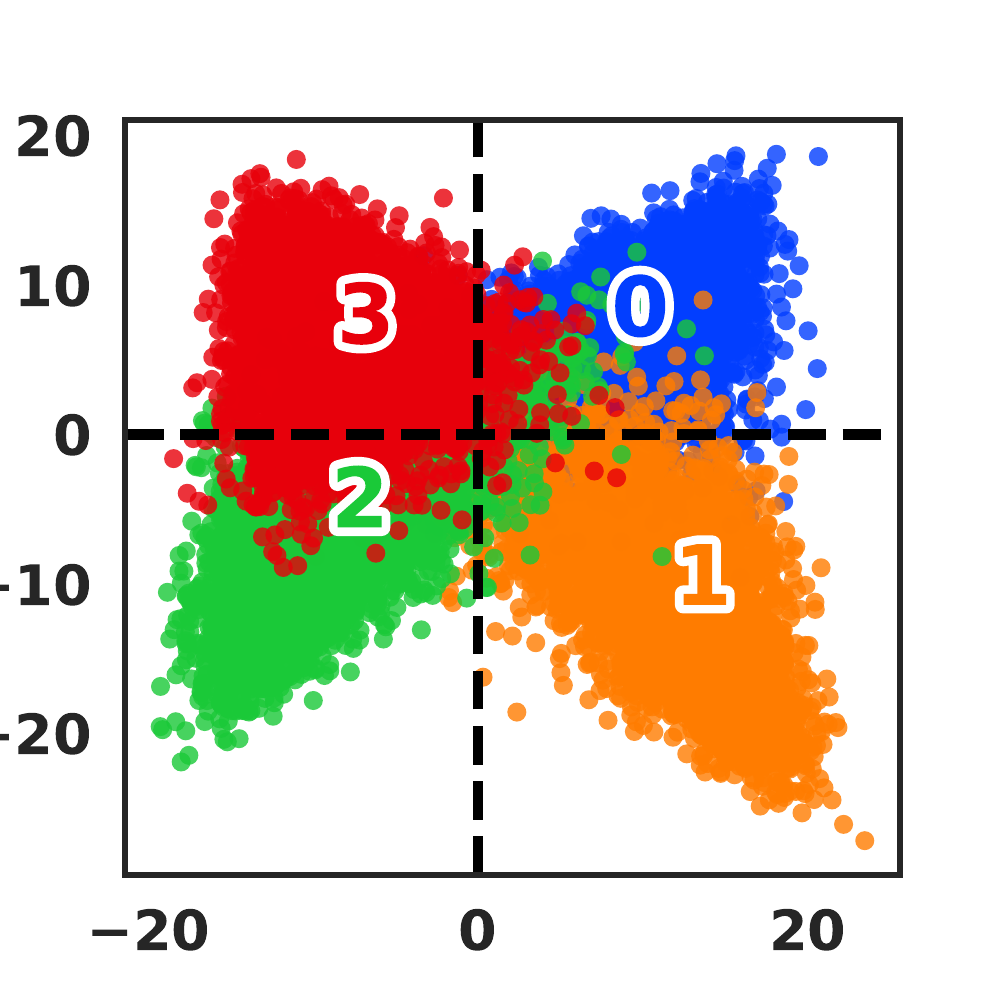}
		\caption{CSQ (mAP: 0.8280)}\label{fig:1c}
	\end{subfigure}
	\quad
	\begin{subfigure}[t]{0.23\textwidth}
		\centering
		\includegraphics[width=1.0 \textwidth]{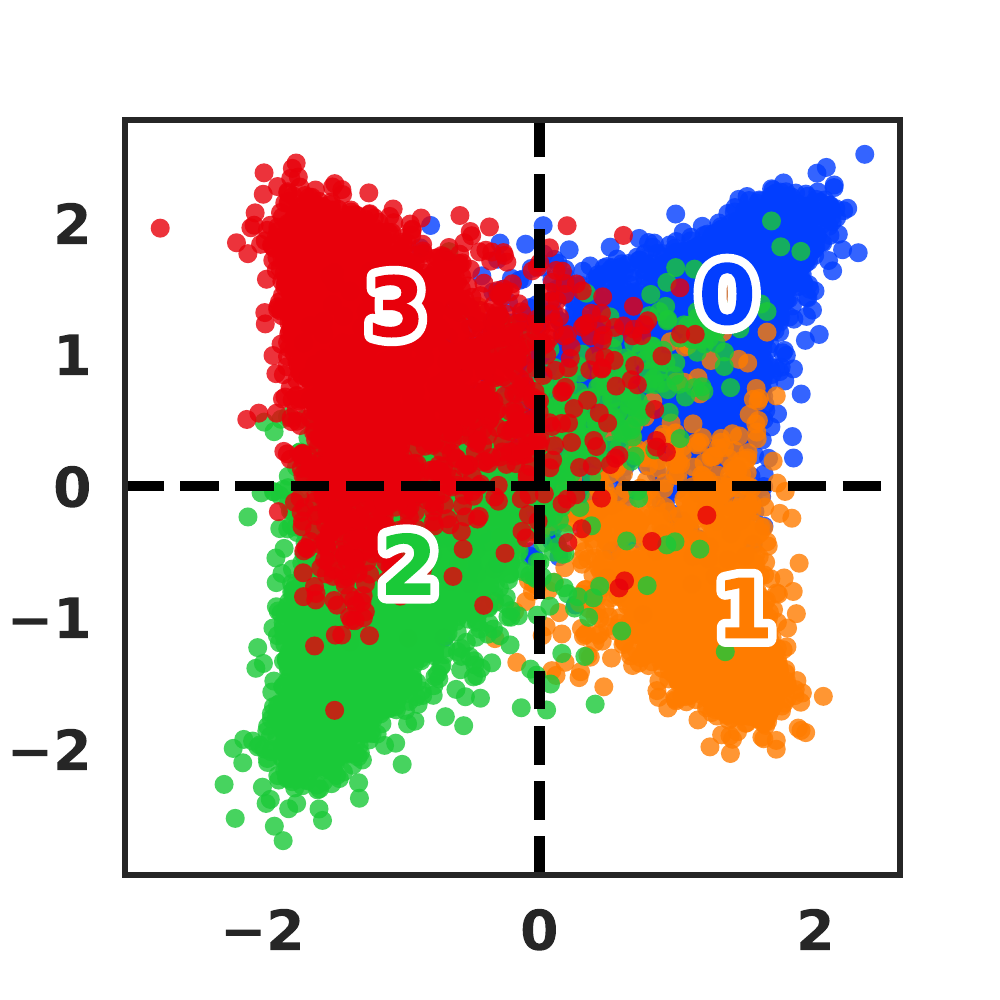}
		\caption{CSQ-C (mAP: 0.8521)}\label{fig:1d}
	\end{subfigure}
	\caption{Visualization of the learned 2-bit hashing space (i.e. learn a 2-bit hash function) using the HashNet and DSDH methods. HashNet-C and DSDH-C replace the quantization in HashNet and DSDH, respectively, with the  proposed HSWD quantization approach.}\label{fig:demo_cifar10}
\end{figure*}
We focus on the general supervised setting of learning a discrete hash function with a neural network. In this setting, a training dataset consists of $N$ images, $X=\{x\}_N$, and a corresponding labeling function $Y$. $Y$ can be point-wise (i.e., assigning labels to each image) or pair-wise (i.e., assigning a similarity score to a pair of images) and is pre-defined. The labeling function essentially describes the semantic similarity of the input. The goal of deep supervised hashing is to learn a discrete hash function ${\bf H}: x \rightarrow c$ that preserves the semantic similarity  of the input. Without loss of generality, we consider the point-wise labeling case and the $\theta$ is the parameter of the hash function. The optimization objective can be defined as:
\begin{equation}
    \min_{\theta} L_s({\bf H}(x), Y(x)), \;\;\; s.t. \; {\bf H}(x) \in \{-1,1\}^m
\end{equation}
where $m$ is the number of bits. In this objective function, $L_s$ preserves the semantic similarity of the input, defined via $Y$. In practice, such a discrete optimization, especially when ${\bf H}$ is a neural network, is intractable. To this end, the discrete ${\bf H}$ is relaxed with a continuous function $h: x \rightarrow [-1, 1]^m$ with a regularization. The general, continuously-relaxed objective function can be written as,
\begin{equation}
    \min_{\theta} L_s(h(x), Y(x)) + \lambda L_q(h(x))
\end{equation}
where the $h(x)$ is usually modeled with a $tanh$ activation function at the output layer. The term $L_q$ denotes the quantization objective, which minimizes the gap between the discrete and continuous solutions.

\subsection{Quantization and Code-balance}

The quantization objective is very important for learning a hash function with a good retrieval performance~\cite{weiss2009spectral,wang2017survey}. This objective typically includes three constraints: quantization error, bit balance, and bit uncorrelation. Low-quantization error alleviates the loss that occurs in assigning ``very'' similar data points to the codes with large Hamming distances.
Bit balance ensures that each bit has the same chance of being $-1$ or $1$, while bit uncorrelation encourages each bit to represent the orthogonal feature of the data. Together, bit-balance and bit-uncorrelation essentially characterize  ``code-balance'', i.e., the condition where training data points are uniformly assigned to each hash code. Code balance is important because it helps minimize the time complexity of both the worst and average cases in retrieval and improve the retrieval quality~\cite{he2011compact}.

Consider Figure~\ref{fig:demo_toy}, where the data, from four classes, is projected on a two-dimensional space (Figure~\ref{fig:demo_toy}(a)). The goal is to learn a two-bit hash function. In Figure~\ref{fig:demo_toy}(c), since the bits are correlated, it is not possible to avoid the collision of data points from different classes in the discrete, hashing space. When the hash function has a high quantization error (Figure~\ref{fig:demo_toy}(d)), some similar data points near the boundary are likely to be assigned to two different hash codes, which results in a higher false-negative rate in retrieval. The optimal hash function, as seen in Figure~\ref{fig:demo_toy}(b) pulls the data points toward the corners (low quantization error) and equally divided the data into four quadrants (code balance) while preserving the semantic similarity of the data (data from different classes occupy  different quadrants).

While the quantization objective is important, existing deep supervised approaches do not effectively learn this objective. Consider the case of learning a two-bit hash function on CIFAR-10 using data from four classes. Figures~\ref{fig:demo_cifar10}(a) and~\ref{fig:demo_cifar10}(c) show the quantization results of two representative supervised hashing methods, HashNet~\cite{cao2017hashnet} and CSQ~\cite{yuan2020central}, respectively. HashNet suffers from high quantization error (unbounded continuous output) and bit correlation (several samples from class 1 and class 3 occupy the hash code $(1,1)$). CSQ is slightly better than HashNet but cannot balance well between preserving semantic similarity and the quantization objective. Our proposed quantization constraint in HashNet-C and CSQ-C, which revises the corresponding quantization objective in HashNet and CSQ, respectively, learns hash codes with better balance and lower quantization error. This leads to improvements in the retrieval performance (from $0.7208$ and $0.8280$ to $0.8500$ and $0.8521$ for HashNet and CSQ, respectively).

In the next section, we will describe the proposed quantization objective, i.e., a single distributional-distance minimization between the learned continuous hash codes and a fixed, uniform discrete distribution. This approach can be easily adapted to existing methods and facilitates these methods to learn better hash functions. Our approach reduces the number of quantization hyperparameters to one, significantly reducing the training process.

\subsection{Distributional Quantization Distance}
As discussed, following the uniform discrete distribution (Figure~\ref{fig:demo_toy}(b)) results in better coding balance and low quantization error. Under this distribution (denoted as $B$), a sample $b \in \{-1,1\}^m$  can be drawn as follows: for each dimension, independently and randomly samples a value of $-1$ or $1$ with equal probability.  We can see that the samples from $B$ exhibit coding balance: each bit is balanced because its value has an equal chance of being $-1$ or $1$, and the bits are uncorrelated since each bit is sampled independently. We propose to minimize the discrepancy between the learned hash distribution and $B$ to improve the coding balance in the learned distribution. This approach also reduces the quantization error since the learned continuous codes will gradually become binary. Formally, we minimize the following quantization objective:
\begin{equation}
    L_q(h(X)) = \mathcal{D}(h(X) || B)
\end{equation}
where  $\mathcal{D}$ denotes the distributional distance function. Note that the constraint is applied on the output space of $h$ (thus, $h(X)$), instead on an individual sample ($h(x)$). Nonetheless, minimizing $\mathcal{D}$ is a non-trivial task, especially when the hashing space is high dimensional and the distribution density of the output space that we are learning cannot be estimated. Furthermore, for a choice of a distribution distance, its estimation should be efficient to compute to be used in practical applications.

Some well-known divergences, such as Kullback–Leibler (KL) or Jensen-Shannon (JS) divergence, generally require the   ability to estimate the learned output distribution of $h(x)$. While we can use an additional network (i.e., similar to the discriminators in GANs~\cite{goodfellow2014generative}) to estimate KL (or JS), learning the neural network with the minimax optimization is computationally expensive. KL and JS are also not well-defined for distributions with non-overlapping supports~\cite{arjovsky2017towards}. To remedy this issue, we first consider the Wasserstein-2 distance. To improve the computation efficiency of the Wasserstein-2 distance, we propose a Sliced-based Wasserstein-2 distance that is more computationally efficient and effective to estimate. 
The quantization objective $\mathcal{D}$ using the Wasserstein-2 distance can  be formulated as follows:
\begin{equation}
    \mathcal{D}(\mu, \nu) =
    \left(\inf_{\gamma \in \Pi(\mu,\nu)} \int_{(z,b) \sim \gamma} p(z,b) ||z-b||_2 dz db \right)^{1/2}
    \label{eqn:reg_primal}
\end{equation}
where  $\Pi(\mu, \nu)$ is the set of all transportation plans $\gamma$ such that their marginal distributions are $\mu$ and $\nu$. 
$\mu$ and $\nu$ define the output distribution of the continuous hash function $h$ and the distribution $B$, respectively. Computing the infimum in Equation~(\ref{eqn:reg_primal}) is difficult since the distribution induced by $z=h(x)$ is not fixed or unknown while employing the Kantorovich-Rubinstein duality to optimize Equation~\eqref{eqn:reg_primal} requires modeling a Lipschitz function using an additional neural network~\cite{arjovsky2017wasserstein}. Fortunately, Wasserstein distance has an elegant yet closed-form solution for one-dimensional continuous measures. Denoting $q_{\mu}$ and $q_{\nu}$ as the density functions of $\mu$ and $\nu$, respectively, the Wasserstein-2 distance between one-dimensional measures $\mu$ and $\nu$ is given by:
\begin{align}
    \mathcal{W}(\mu,\nu) = \left(\int_{0}^{1} ||F_{\mu}^{-1}(w)-F_{\nu}^{-1}(w)||_2 dw\right)^{1/2}
\end{align}
where $F_{\mu}(w)=\int_{\infty}^w q_{\mu}(\rho) d\rho$ and $F_{\nu}(w)=\int_{\infty}^w q_{\nu}(\rho) d\rho$ are the cumulative distribution functions. Inspired by the efficiency of estimating the one-dimensional Wasserstein-2 distance, we propose to find a family of one-dimensional representations, e.g., through the linear projections, and approximate the Wasserstein-2 distance as a function of these one-dimensional marginals:
\begin{align}
    \mathcal{D}(h(X), B) \approx \left( \frac{1}{L} \sum_{l=1}^{L} \mathcal{W}(\omega_l^T h(X), \omega_l^T B)\right)^{1/2}
\end{align}
where $\omega_l^T h(X)$ and $\omega_l^T B$ are the projections of the output samples of $h$ and samples from $B$ onto a one-dimensional direction defined by $\omega_l$ (a slice). Typically, $\omega_l$ is drawn from a uniform distribution on the unit sphere. This is also known as the Sliced-Wasserstein distance (SWD)~\cite{deshpande2018generative,kolouri2019generalized}. While this approach has successful applications in a variety of tasks~\cite{deshpande2018generative,kolouri2019generalized}, the random nature of the slices could lead to several non-informative directions where the sliced distances along  these directions are close to 0. Consequently, a large number $L$ of random directions is needed to approximate the sliced-Wasserstein distance, which increases the computational complexity of the estimation.

\begin{figure}[b!]

\vspace{-0.08in}

\mbox{\hspace{-0.2in}
	\begin{subfigure}[t]{1.75in}
		\centering
		\includegraphics[width=1.0\textwidth]{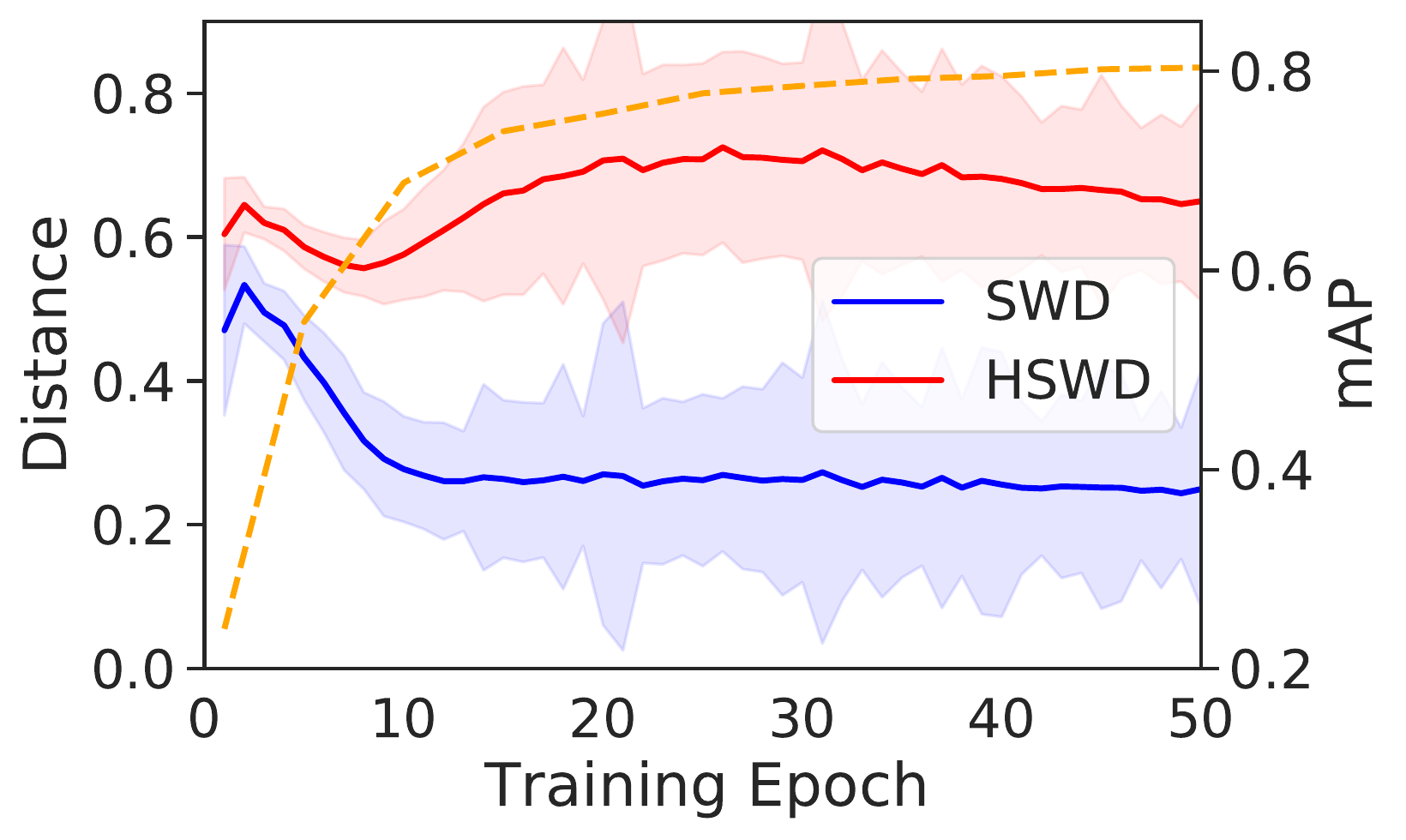}
		\caption{\textbf{DSDH}: SWD Train}\label{}
	\end{subfigure}
	\begin{subfigure}[t]{1.75in}
		\includegraphics[width=1.0\textwidth]{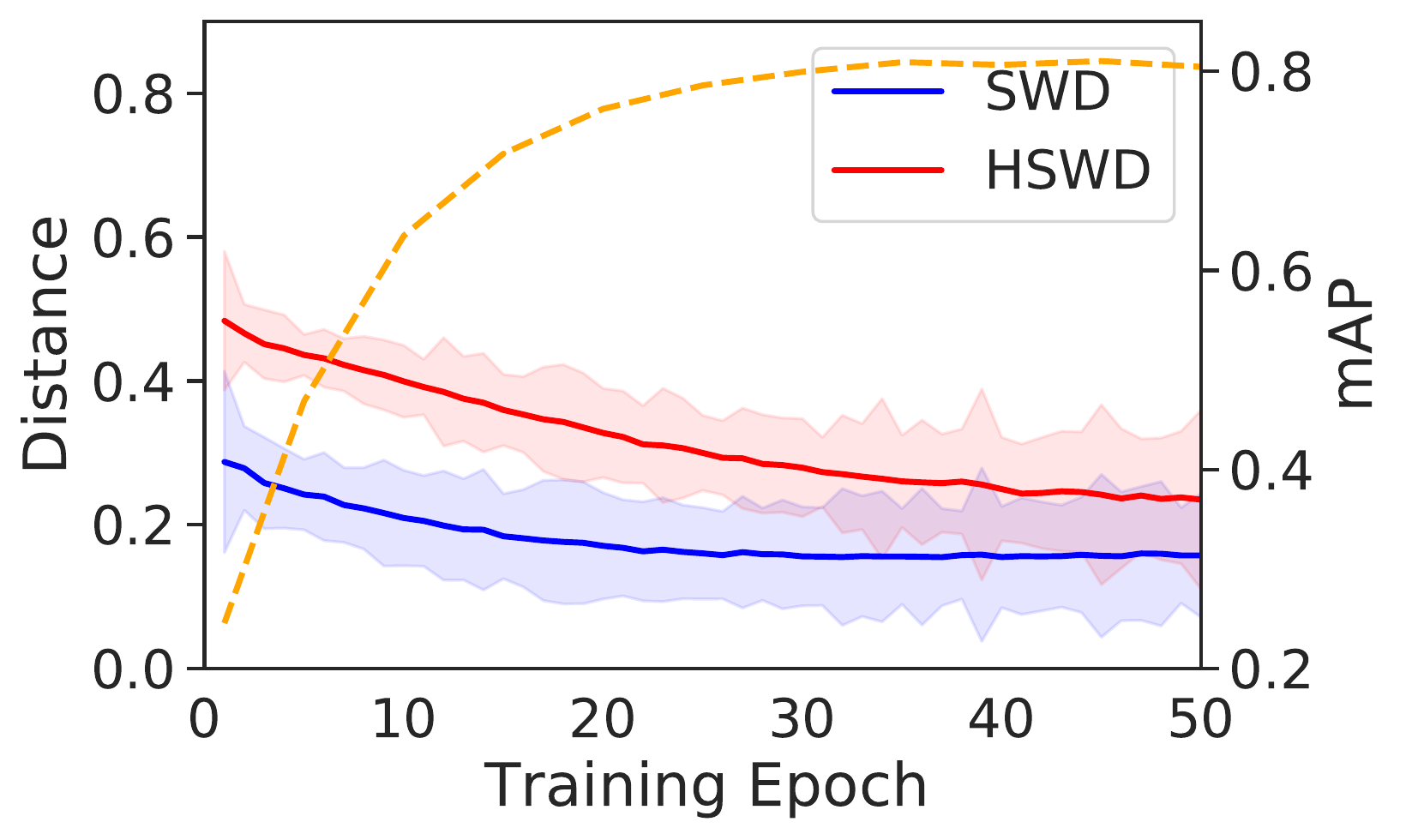}
		\caption{\textbf{DSDH}: HSWD Train}\label{}
	\end{subfigure}
}

\mbox{\hspace{-0.2in}
	\begin{subfigure}[t]{1.75in}
		\includegraphics[width=1.0\textwidth]{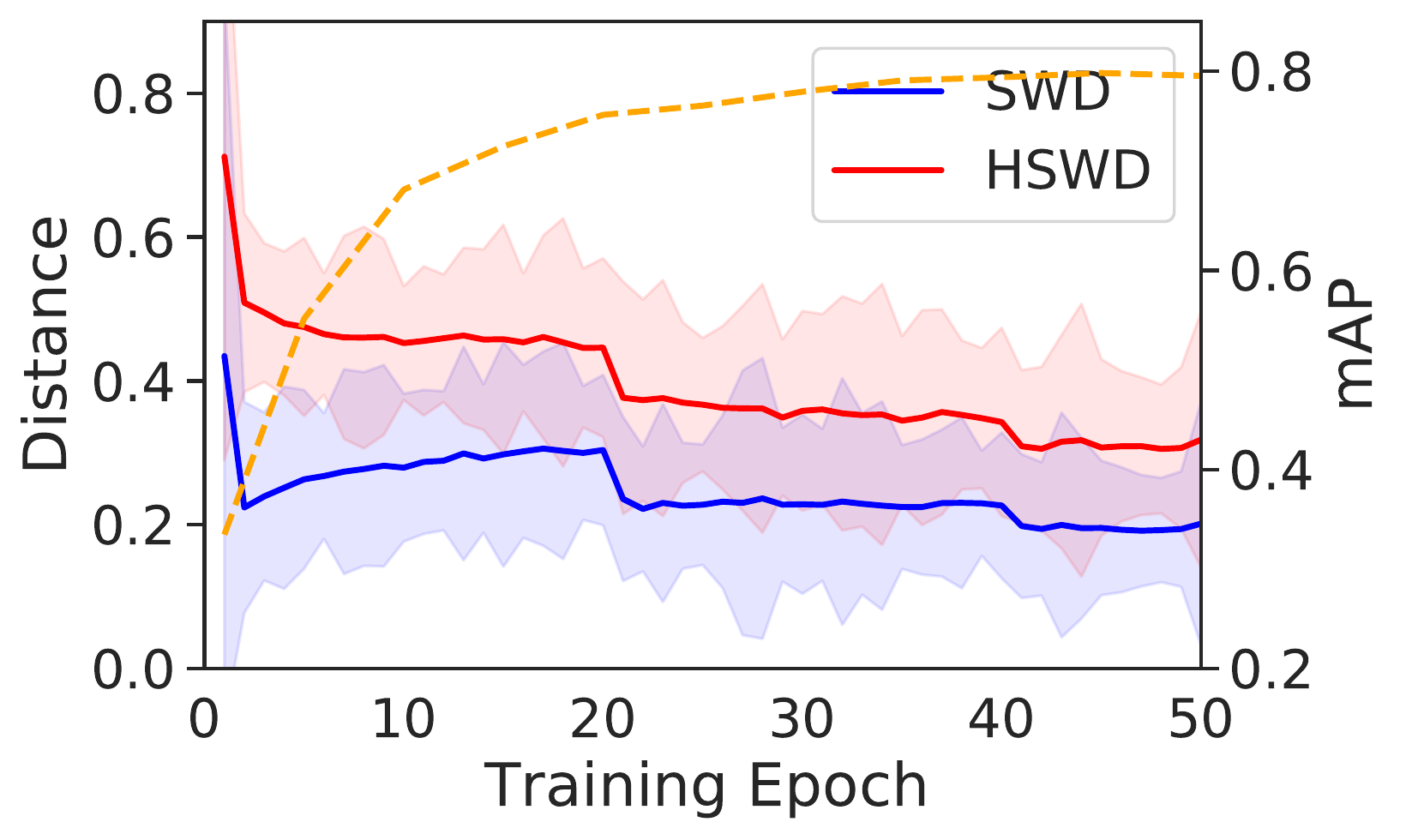}
		\caption{\textbf{HashNet}: SWD Train}\label{}
	\end{subfigure}
	\begin{subfigure}[t]{1.75in}
		\includegraphics[width=1.0\textwidth]{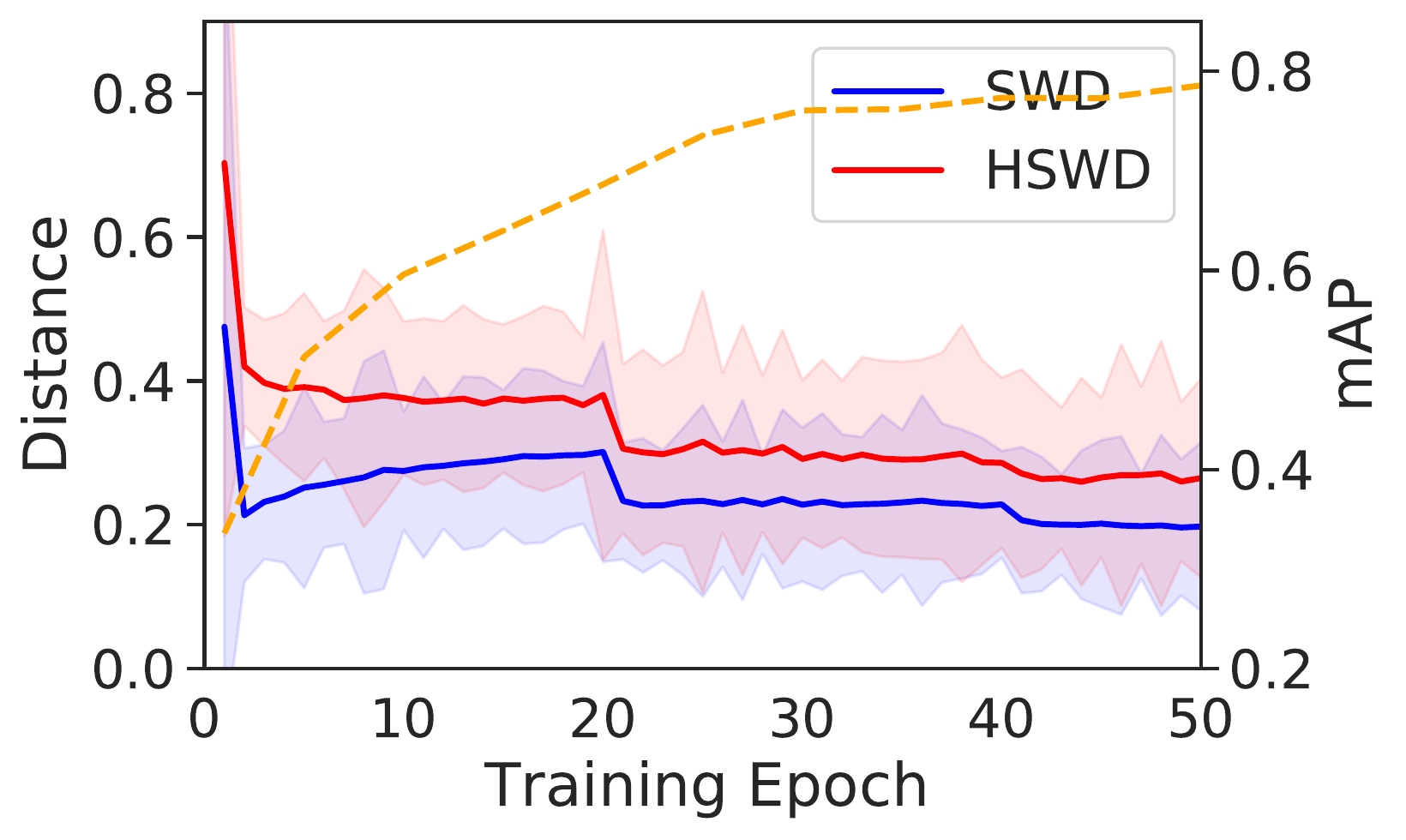}
		\caption{\textbf{HashNet}: HSWD Train}\label{}
	\end{subfigure}
}
	\caption{Estimated SWD and HSWD during training the 32-bit hash functions on CIFAR-10 dataset. For each method (DSDH~\cite{li2017deep} or HashNet~\cite{cao2017hashnet}), SWD or HSWD is used as the quantization objective. The dashed (orange) line denotes the retrieval performance (mAP).}\label{fig:demo_estimate_distances} 
\end{figure}

To remedy this issue, we avoid the random projections and select directions that contain discriminant information of the two data sources. 
We utilize the observation that the objective $L_s$ aims to learn discriminative projections, via $h$, of the data points onto the discrete space. Intuitively, since each output dimension of the hash function describes a discriminative feature of the space, projecting the data into this dimension can capture meaningful separation of the two distributions. Empirically, \py{how to define HSWD?} we can show that the averaged one-dimensional Wasserstein distance along these dimensions, denoted as HSWD, captures better separations of the data than SWD. Specifically, we provide the estimated HSWD and SWD (using 10,000 projections in CIFAR-10) when the model is trained to minimize SWD (Figures~\ref{fig:demo_estimate_distances}(a) and~\ref{fig:demo_estimate_distances}(c)), or HSWD (Figures~\ref{fig:demo_estimate_distances}(b) and~\ref{fig:demo_estimate_distances}(d)) as the quantization objective. As we can observe, HSWD describes better separations of the two distributions (i.e., higher distances). Minimizing SWD does not effectively bring the two distributions closer in some cases (e.g., HSWD increases in Figure~\ref{fig:demo_estimate_distances}(a), indicating the separations even increase in some directions). However, minimizing HSWD effectively forces $h(X)$ to be closer to $B$, as both HSWD and SWD decrease in Figures~\ref{fig:demo_estimate_distances}(b) and~\ref{fig:demo_estimate_distances}(d). More importantly, the number of projections, which is also the number of dimensions of $h(X)$, is fixed and generally small ($m \le 128$ in most applications). The one-dimensional distances along these dimensions are exactly the distances along the coordinate axis of the discrete space. Thus, HSWD is defined, as follows:
\begin{align}
    \mathcal{D}(h(X), B) \approx \left( \frac{1}{m} \sum_{l=1}^{m} [\mathcal{W}(h(X)_{l,:}, B_{l,:})]^2\right)^{1/2}
\end{align}

where $h(X)_{l,:}$ and $B_{l,:}$ are the one-dimensional samples along the dimension $l$ of $h(X)$ and $B$, respectively.
Note that, using the discrete dimensions as projected directions is similar to the works in~\cite{doan2020imagehash} for the unsupervised study. We can show that HSWD is a valid distance.

\begin{theorem}
The proposed distance Sliced Wasserstein calculation  in Theorem 1, denoted as HSWD, is a valid distance function of probability measures in this space.
\end{theorem}

\begin{remark}
The proposed HSWD is similar to the Max Sliced Wasserstein Distance (M-SWD), which has successful applications~\cite{deshpande2019max,kolouri2019generalized}. However, different from M-SWD, HSWD  does not require a separate discriminative network to estimate the distance.
\end{remark}



HSWD enjoys a better computational efficiency than SWD ($O(mN \log(Nd))$ compared to $O(LN \log(Nd))$). In most problems, SWD requires a large number ($L \gg N$) of random directions, typically between 1000 to 10,000, to provide a reliable estimate of the distance~\cite{deshpande2019max,nguyen2020distributional}. In HSWD, the number of directions is fixed to the dimension of the hashing space $m$, which is typically between 16 to 128 for many image hashing applications.

\section{Experiments}\label{sec:experiments}

We present the experimental results to demonstrate the effectiveness of the proposed quantization method over the existing quantization approaches using extensive combinations of representative datasets, deep supervised hashing methods, and hash-function architectures.

\subsection{Datasets Used}

We utilize the following datasets:

\begin{itemize}
    \item \textbf{CIFAR10}: This dataset consists of 60,000 natural images categorized uniformly into 10 labels. We randomly select 100 and 500 images from each label for the query set and the training set, respectively. The remaining images form the retrieval set. Hence, there are 1000, 5,000, and 54,000 images in the query, training, and retrieval sets, respectively.

    \item \textbf{NUS-WIDE}: This dataset contains 269,648 images, each of which belongs to at least one of 21 concepts. We randomly select 5,000 images for the query set, with the remaining images used as the retrieval set. 10,000 images randomly selected in the retrieval set are used for training.

    \item \textbf{COCO}: This dataset contains 123,287 images, labeled with at least 1 out of 80 semantic concepts. Similarly, the query set is randomly constructed with 5,000 images, with the remaining images used as the retrieval set (10,000 randomly selected images in this set are used for training).
\end{itemize}

\subsection{Evaluation Metrics}

For evaluating the performance of the proposed model, we follow the standard evaluation mechanism that is widely accepted for the problem of image hashing
- the \textbf{precision@R} (\textbf{P@R}) and \textbf{mean average precision} (\textbf{mAP}).

\begin{table*}[!htp]
\newcommand{\improv}[2]{#1/\textbf{\color{blue}#2\%}}
\newcommand{\notimprov}[2]{#1/\textit{\color{black}#2\%}}
\setlength{\tabcolsep}{3pt}
\footnotesize
    \centering
    \caption{mAP for different numbers of bits on the three image datasets. The blue value (in bold) along the mAP value of each proposed approach shows the relative improvement over the original algorithm, while the italicized value indicates no improvement.}
    \vspace{-5pt}
    \begin{tabular}{l|lll|lll|lll}
        \toprule

        \multirow{2}{*}{Method} & \multicolumn{3}{|c}{CIFAR-10} & \multicolumn{3}{|c}{NUS-WIDE} & \multicolumn{3}{|c}{COCO} \\
        \cline{2-10}
        & 16 bits & 32 bits & 64 bits & 16 bits & 32 bits & 64 bits & 16 bits & 32 bits & 64 bits  \\

        \hline

        DSDH~\cite{li2017deep} & 0.7909 & 0.8072 & 0.8278 & 0.8270 & 0.8455 & 0.8640 & 0.7331 & 0.7853 & 0.8074  \\
        DSDH-S & \improv{0.8187}{3.5} & \improv{0.8439}{4.6} & \improv{0.8517}{2.9} & \improv{0.8282}{0.1} & \improv{0.8461}{0.1} & \improv{0.8712}{0.8} & \notimprov{0.7330}{0.0} & \improv{0.8030}{2.3} & \improv{0.8404}{4.1} \\

        DSDH-C & \improv{0.8531}{7.9} & \improv{0.8620}{6.8} & \improv{0.8658}{4.6}  & \improv{0.8433}{2.0} & \improv{0.8631}{2.1} & \improv{0.8749}{1.3} & \improv{0.7424}{1.3} & \improv{0.8032}{2.3} & \improv{0.8408}{4.1} \\

        \hline

        HashNet~\cite{cao2017hashnet} & 0.6922 & 0.8311 & 0.8566 & 0.7728 & 0.8336 & 0.8654 & 0.6899 & 0.7666 & 0.8098 \\
        HashNet-S & \improv{0.8131}{17} & \improv{0.8573}{3.2} & \improv{0.8749}{2.1} & \improv{0.8062}{4.3} & \improv{0.8438}{1.2} & \improv{0.8713}{0.7} & \improv{0.7215}{4.6} & \improv{0.7764}{1.3} & \improv{0.8189}{1.1} \\

        HashNet-C & \improv{0.7939}{14} & \improv{0.8467}{1.9} & \improv{0.8691}{1.5} & \improv{0.8002}{3.5} & \improv{0.8437}{1.2} & \improv{0.8791}{1.6} & \improv{0.7202}{4.4} & \improv{0.7789}{1.6} & \improv{0.8202}{1.3} \\

        \hline

        GreedyHash~\cite{su2018greedy} & 0.8223 & 0.8474 & 0.8646 & 0.7802 & 0.8081 & 0.8328 & 0.6533 & 0.7219 & 0.7561 \\
        GreedyHash-S & \improv{0.8280}{0.7} & \improv{0.8497}{0.3} & \improv{0.8653}{0.1} & \improv{0.7815}{0.1} & \notimprov{0.8083}{0.0} & \improv{0.8390}{0.7} & \improv{0.6668}{2.1} & \improv{0.7291}{1.0} & \improv{0.7618}{0.8} \\
        GreedyHash-C & \improv{0.8375}{1.9} & \improv{0.8536}{0.7} & \improv{0.8722}{0.9} & \improv{0.7890}{1.1} & \improv{0.8179}{1.2} & \improv{0.8477}{1.8} & \improv{0.6637}{1.6} & \improv{0.7299}{1.1} & \improv{0.7712}{2.0} \\

        \hline

        DCH~\cite{cao2018deep} & 0.8302 & 0.8432 & 0.8558 &  0.8015 & 0.8061 & 0.8040 & 0.7578 & 0.7792 & 0.7723 \\
        DCH-S & \improv{0.8372}{0.8} & \improv{0.8515}{1.0} & \improv{0.8602}{0.5} & \improv{0.8058}{0.5} & \improv{0.8079}{0.2} & \improv{0.8067}{0.3} & \improv{0.7657}{1.1} & \improv{0.7831}{0.5} & \improv{0.7803}{1.0} \\
        DCH-C & \improv{0.8446}{1.7} & \improv{0.8596}{1.9} & \improv{0.8711}{1.8} &  \improv{0.8159}{1.8} & \improv{0.8145}{1.0} & \improv{0.8155}{1.4} & \improv{0.7702}{1.6} & \improv{0.7892}{1.3} & \improv{0.7807}{1.1} \\        

        \hline

        CSQ~\cite{yuan2020central} & 0.8069 & 0.8291 & 0.8366 & 0.7992 & 0.8384 & 0.8596 & 0.6783 & 0.7550 & 0.8146 \\
        CSQ-S & \improv{0.8401}{4.1} & \improv{0.8555}{3.2} & \improv{0.8554}{2.3} & \improv{0.8044}{0.7} & \improv{0.8495}{1.3}  & \improv{0.8626}{0.4}  & \improv{0.7036}{3.7} & \improv{0.7765}{2.8} & \improv{0.8234}{1.0}  \\
        CSQ-C & \improv{0.8457}{4.8} & \improv{0.8558}{3.2} & \improv{0.8652}{3.4} & \improv{0.8054}{0.8} & \improv{0.8511}{1.5} & \improv{0.8701}{1.2} & \improv{0.6989}{3.0} & \improv{0.7752}{2.7} & \improv{0.8255}{1.3} \\
        
        \hline

        DBDH~\cite{zheng2020deep} & 0.7660 & 0.8223 & 0.8492 & 0.8305 & 0.8552 & 0.8666 & 0.7202 & 0.7826 & 0.8042 \\
        DBDH-S & \improv{0.8458}{10} & \improv{0.8587}{4.4} & \improv{0.8603}{1.3} & \improv{0.8387}{1.0} & \improv{0.8577}{0.3} & \improv{0.8680}{1.8} & \improv{0.7461}{2.2} & \improv{0.7996}{3.7} & \improv{0.8336}{4.3} \\

        DBDH-C & \improv{0.8466}{10} & \improv{0.8593}{4.5} & \improv{0.8668}{2.1} & \improv{0.8395}{1.1} & \improv{0.8633}{0.9} & \improv{0.8760}{1.1} & \improv{0.7389}{2.6} & \improv{0.7889}{0.8} & \improv{0.8308}{3.9} \\

        \bottomrule
    \end{tabular}
    \label{tab:result_map}
    \vspace{-10pt}
\end{table*}

\begin{table}[!htp]
\newcommand{\improv}[2]{#1/\textbf{\color{blue}#2\%}}
\newcommand{\notimprov}[2]{#1/\textit{\color{black}#2\%}}
\setlength{\tabcolsep}{2pt}
\footnotesize
    \centering
    \caption{P@1000 for 16 and 32 bits on CIFAR-10 and NUS-WIDE. The blue value (in bold) along the P@1000 value of each proposed approach shows the relative improvement over the original algorithm, while the italicized value indicates no improvement.}
    \vspace{-5pt}
\mbox{\hspace{-0.05in}
    \begin{tabular}{l|ll|ll}
        \toprule

        \multirow{2}{*}{Method} & \multicolumn{2}{|c}{CIFAR-10} & \multicolumn{2}{|c}{NUS-WIDE} \\
        \cline{2-5}
        & 16 bits & 32 bits & 16 bits & 32 bits \\

        \hline

DSDH	& 0.8252 & 0.8406 & 0.8117 & 0.8294	\\
DSDH-S	& \improv{0.8526}{3.3} & 	\improv{0.8543}{1.6} & 	\improv{0.8162}{0.6} & 	\improv{0.8312}{0.2} \\
DSDH-C & 	\improv{0.8645}{4.8} & 	\improv{0.8739}{4.0} & 	\improv{0.8195}{1.0} & 	\improv{0.8391}{1.2} \\

        \hline

HashNet	& 0.6193 & 0.8613 & 0.7581 & 0.8158 \\
HashNet-S	& \improv{0.8470}{36.8} & 	\improv{0.8755}{1.7} & 	\improv{0.7743}{2.1} & 	\improv{0.8199}{0.5} \\
HashNet-C	& \improv{0.7698}{24.3} & 	\improv{0.8715}{1.2} & 	\notimprov{0.7456}{-1.7} & 	\notimprov{0.8078}{-1.0} \\

        \hline

GreedyHash	& 0.8561 & 0.8616 & 0.7601 & 0.8009 \\
GreedyHash-S	& \improv{0.8583}{0.3} & 	\improv{0.8656}{0.5} & 	\improv{0.7657}{0.7} & 	\notimprov{0.7973}{-0.5} \\
GreedyHash-C	& \notimprov{0.8517}{-0.5} & 	\improv{0.8700}{1.0} & 	\improv{0.7630}{0.4} & 	\notimprov{0.7931}{-1.0} \\

        \hline

DCH	& 0.8621 & 0.8568 & 0.7843 & 0.7898 \\
DCH-S	& \notimprov{0.8622}{0.0} & 	\improv{0.8761}{2.3} & 	\notimprov{0.7846}{0.0} & 	\improv{0.7923}{0.3} \\
DCH-C	& \improv{0.8654}{0.4} & 	\improv{0.8635}{0.8} & \improv{0.7893}{0.6} & 	\improv{0.7914}{0.2} \\

        \hline

CSQ	& 0.8510 & 0.8571 & 0.7903 & 0.8285 \\
CSQ-S	& \improv{0.8661}{1.8} & 	\improv{0.8732}{1.9} & 	\improv{0.8034}{1.7} & 	\improv{0.8318}{0.4} \\
CSQ-C	& \improv{0.8670}{1.9} & 	\improv{0.8688}{1.4} & 	\improv{0.8007}{1.3} & 	\improv{0.8353}{0.8} \\

        \hline

DBDH	& 0.8440 & 0.8421 & 0.8122 & 0.8323 \\
DBDH-S & 	\improv{0.8626}{2.2} & 	\improv{0.8675}{3.0} & \improv{0.8177}{0.7} & 	\improv{0.8388}{0.8} \\
DBDH-C & 	\improv{0.8658}{2.6} & 	\improv{0.8731}{3.7} &  	\improv{0.8135}{0.1} & 	\improv{0.8380}{0.7} \\

        \bottomrule
    \end{tabular}
    }
    \label{tab:result_precision_alexnet}
\end{table}
\subsection{Comparison Methods}

We compare the performance of the proposed quantization approach in various representative deep supervised image hashing methods: \textbf{DSDH}~\cite{li2017deep}, \textbf{HashNet}~\cite{cao2017hashnet}, \textbf{GreedyHash}~\cite{su2018greedy}, \textbf{DCH}~\cite{cao2018deep}, \textbf{CSQ}~\cite{yuan2020central}, and \textbf{DBDH}~\cite{zheng2020deep}. Note that the performance comparisons between the methods are beyond the scope of this paper and we only focus on the performance improvement when the proposed quantization approach is incorporated in these methods.

For each method, report the retrieval performance using the original quantization proposal and our quantization approach using SWD (using the suffix -S) and HSWD (using the suffix -C). For example, for the CSQ method, we report CSQ (original algorithm), CSQ-S (CSQ with SWD quantization), and CSQ-C (CSQ with HSWD quantization).

\subsection{Implementation Details}
For a fair comparison, we use the same underlying deep hash function (e.g., VGG11) for the original algorithm and those with the proposed quantization approaches. We perform a hyperparameter selection step for each method and report the average performances for the best configuration across multiple runs with different random initializations. 

\subsection{Retrieval Performance}

In this section, we measure the performance of the proposed quantization approach and the original quantization proposal for each selected deep supervised hashing method. Table~\ref{tab:result_map} shows the mAP results in our experiments for learning the 16-bit, 32-bit, and 64-bit hash functions with the VGG11 backbone~\cite{simonyan2014very}. We also report the P@1000 results for learning the 16-bit and 32-bit hash functions with the AlexNet backbone~\cite{krizhevsky2012imagenet} in Table~\ref{tab:result_precision_alexnet}. For each of the proposed quantization approaches, we report the relative improvement besides the mAP or P@1000 values. We highlight the statistically-significant relative improvements in blue (and in bold values), while the values are italic when there are not statistically-significant improvements.

As we can observe in Tables~\ref{tab:result_map}~and~\ref{tab:result_precision_alexnet}, the proposed quantization approaches achieve significant improvement in the retrieval results across different supervised hashing methods. Specifically, the SWD variants result in up to over 10\% improvement in CIFAR-10, 4\% in NUS-WIDE, and almost 5\% in COCO. Similarly, the HSWD variants also consistently improve the retrieval results in these supervised hashing methods. As mentioned previously, the primary advantage of HSWD over SWD is its computational efficiency. We conjecture that the superior performance of the proposed approach is due to the following reasons:

\begin{itemize}
    \item The quantization constraints in the existing deep supervised hashing methods are less effective to minimize the gaps between the relaxed hash function $h(x)$ and the optimal, discrete hash function $H(x)$. The observation in Figure~\ref{fig:demo_cifar10} also confirms our discussion.

    \item The proposed quantization approaches can help learn better-quantized hash functions. Specifically, it is more effective to optimize the combination of the similarity preserving loss $L_s$ and $L_q$ when $L_q$ is one of our proposed approaches.

    \item Finally, low-quantization error and code balance are crucial in improving the retrieval performance in the existing deep supervised hashing methods. Additional empirical results for this relationship are presented in Section~\ref{sec:quantization_vs_retrieval}. 
\end{itemize}

\subsection{Computational Efficiency}

We compare the training time per epoch of the proposed quantization approaches when learning 64-bit hash functions. For each method, we capture the average running time (in seconds) of the original algorithm and the algorithm with SWD- and HSWD-quantizations. For SWD, we report the corresponding running time of the number of projections that results in the optimal performance. We observe that the optimal number of projections ranges from 1000 to 10,000. Note that for HSWD, the number of projections is fixed and equals the size of the hash codes. 

Table~\ref{tab:results_running_time} report\py{s} the average running times across different deep hashing methods. We can observe that HSWD is more computationally efficient than SWD because of its fewer projections and  the omission of the matrix-multiplication operation that projects the data points into random directions. Compared to the original quantizations, HSWD quantization is also faster since each original quantization (in $L_q$) usually comprises of several losses.

\begin{table}[t]
    \newcommand{\improv}[2]{#1/\textbf{\color{blue}#2\%}}
    \centering
    \caption{Averaged running time per epoch across different supervised hashing methods (in seconds). The blue values are the relative running-time decreases.}
    \begin{tabular}{l|c|c|c}
        \toprule
        Dataset & Original & SWD & HSWD  \\ \hline
        CIFAR-10 & 19.4 & 24.2 & \improv{17.1}{40}  \\
        NUS-WIDE & 58.3 & 71.2 & \improv{50.1}{41} \\
        COCO & 55.6 & 68.1 & \improv{49.5}{37} \\
        \bottomrule
    \end{tabular}
    \label{tab:results_running_time}
\end{table}

\subsection{Qualitative Analysis}

\subsubsection{Hash Code Visualization}

\begin{figure}[!h]
	\centering
	\begin{subfigure}[t]{0.135\textwidth}
		\centering
		\includegraphics[width=1.0 \textwidth]{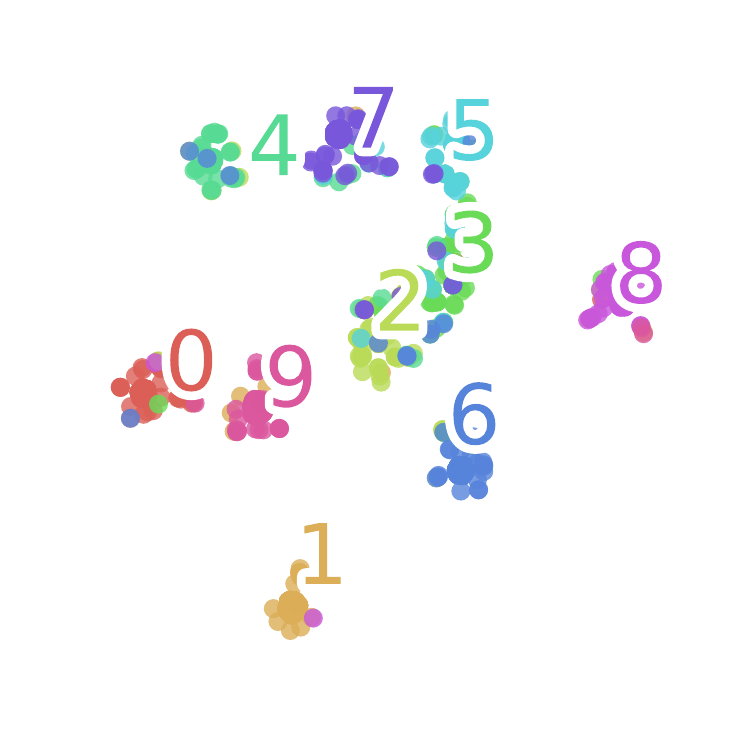}
		\caption{CSQ}\label{fig:4a}
	\end{subfigure}
	\quad
	\begin{subfigure}[t]{0.135\textwidth}
		\centering
		\includegraphics[width=1.0 \textwidth]{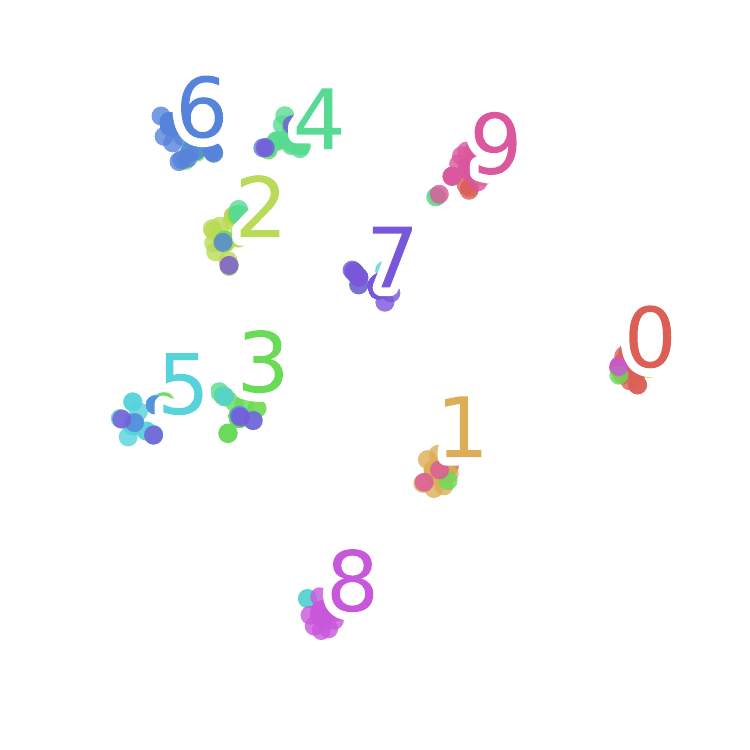}
		\caption{CSQ-S}\label{fig:4b}
	\end{subfigure}
	\quad
	\begin{subfigure}[t]{0.135\textwidth}
		\centering
		\includegraphics[width=1.0 \textwidth]{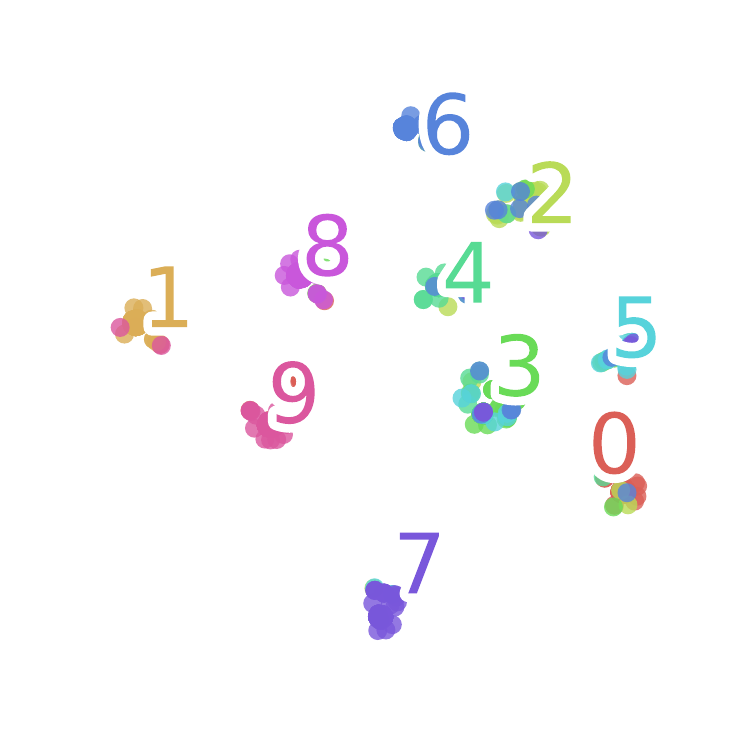}
		\caption{CSQ-C}\label{fig:4c}
	\end{subfigure}
	\vfill
	\begin{subfigure}[t]{0.135\textwidth}
		\centering
		\includegraphics[width=1.0 \textwidth]{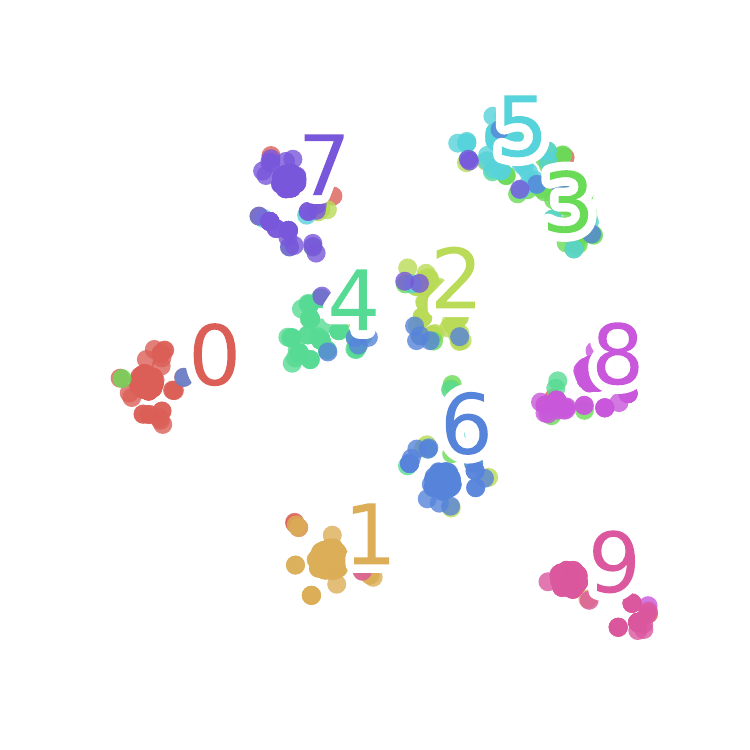}
		\caption{DCH}\label{fig:4d}
	\end{subfigure}
	\quad
	\begin{subfigure}[t]{0.135\textwidth}
		\centering
		\includegraphics[width=1.0 \textwidth]{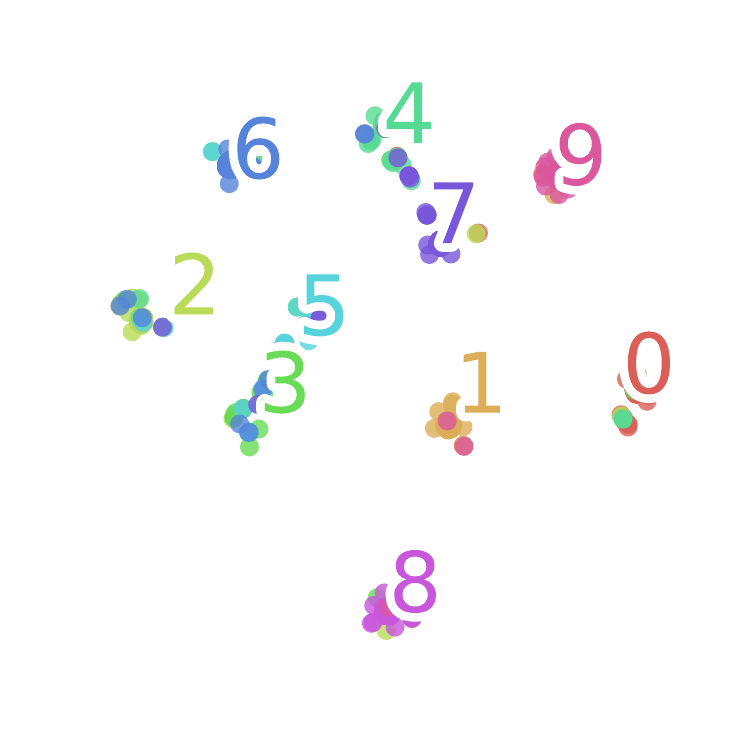}
		\caption{DCH-S}\label{fig:4e}
	\end{subfigure}
	\quad
	\begin{subfigure}[t]{0.135\textwidth}
		\centering
		\includegraphics[width=1.0 \textwidth]{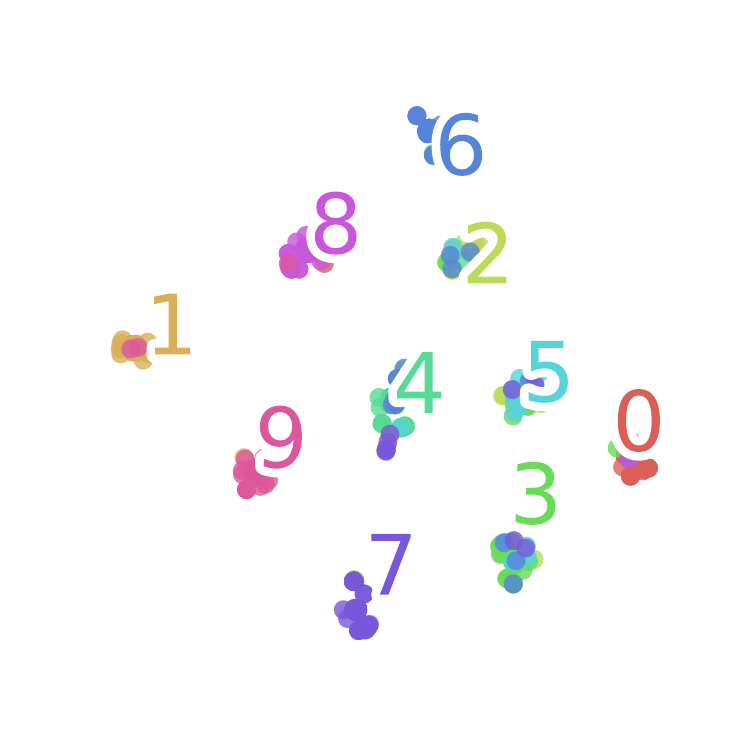}
		\caption{DCH-C}\label{fig:4f}
	\end{subfigure}
	\caption{Two-dimensional t-SNE visualizations of the quantized 16-bit hash codes learned by different quantization approaches in CSQ and DCH on CIFAR-10.}\label{fig:results_tsne}
\end{figure}

We present the visualization of the hash codes generated by the learned hash functions for the different quantization approaches, using the CSQ and DCH methods, in Figure~\ref{fig:results_tsne}. We project the hash codes into 2-dimensional embeddings using the t-SNE method. As Figure~\ref{fig:results_tsne} displays, the learned hash codes of the methods with SWD and HSWD quantizations exhibit better inter-cluster separation and intra-cluster closeness, which makes their retrieval performance superior. For example, for the CSQ's result in Figure~\ref{fig:results_tsne}(a), we can observe that the samples from classes 2 and 3 are closer to each other. For CSQ-S and CSQ-C, the samples from these two classes are more separated.

\subsubsection{Analysis of Effective Quantization and Performance Improvement}\label{sec:quantization_vs_retrieval}
\begin{figure}[!htpb]
	\centering
\mbox
{\hspace{-0.25in}	
	\begin{subfigure}[t]{0.26\textwidth}
		\centering
		\includegraphics[width=1.0 \textwidth]{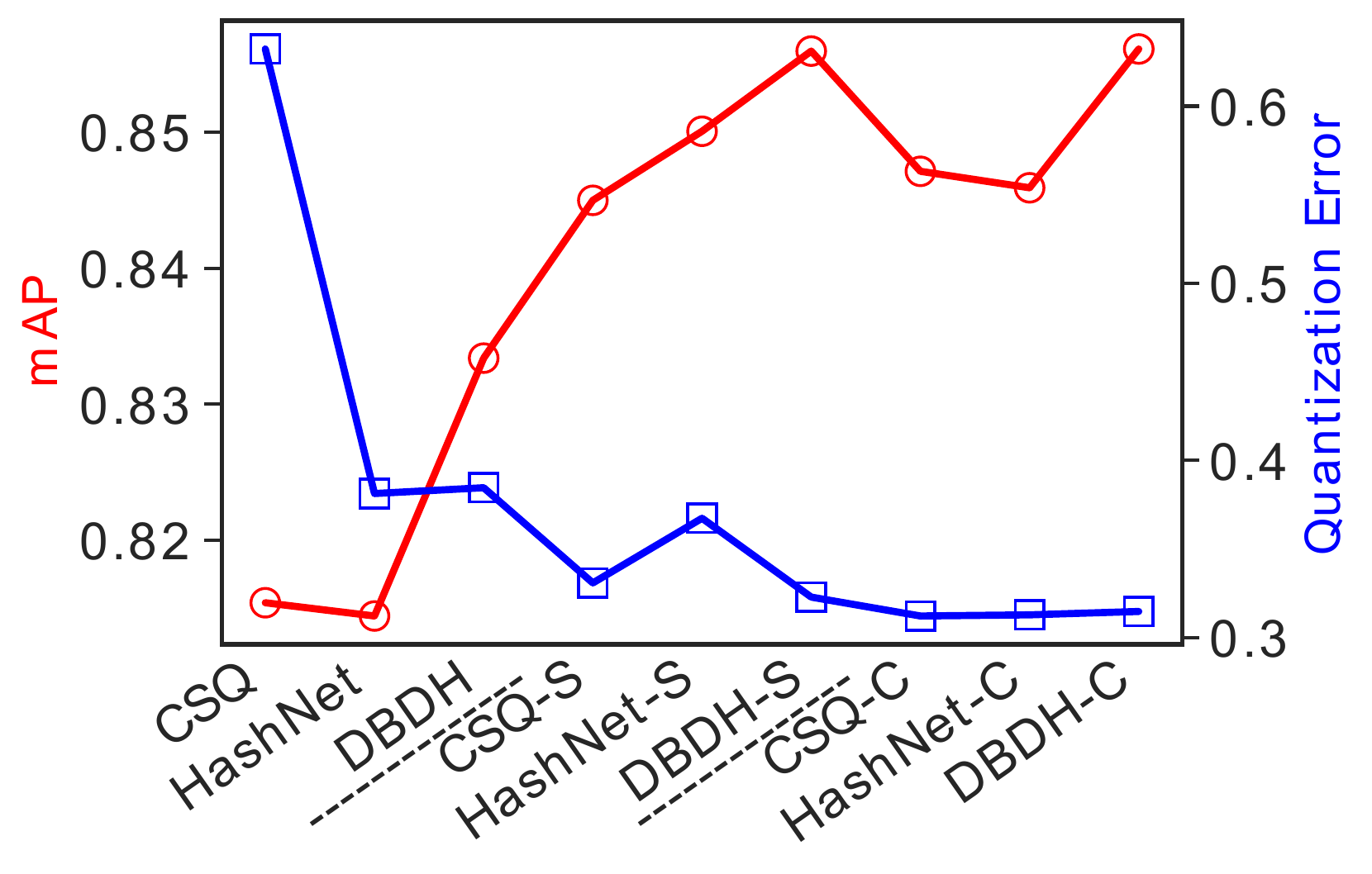}
		\caption{Quantization Error}\label{fig:1a}
	\end{subfigure}

	\begin{subfigure}[t]{0.26\textwidth}
		\centering
		\raisebox{2.6mm}[0pt][0pt]
		{%
        \makebox[\textwidth][c]{\includegraphics[width=1.0 \textwidth]{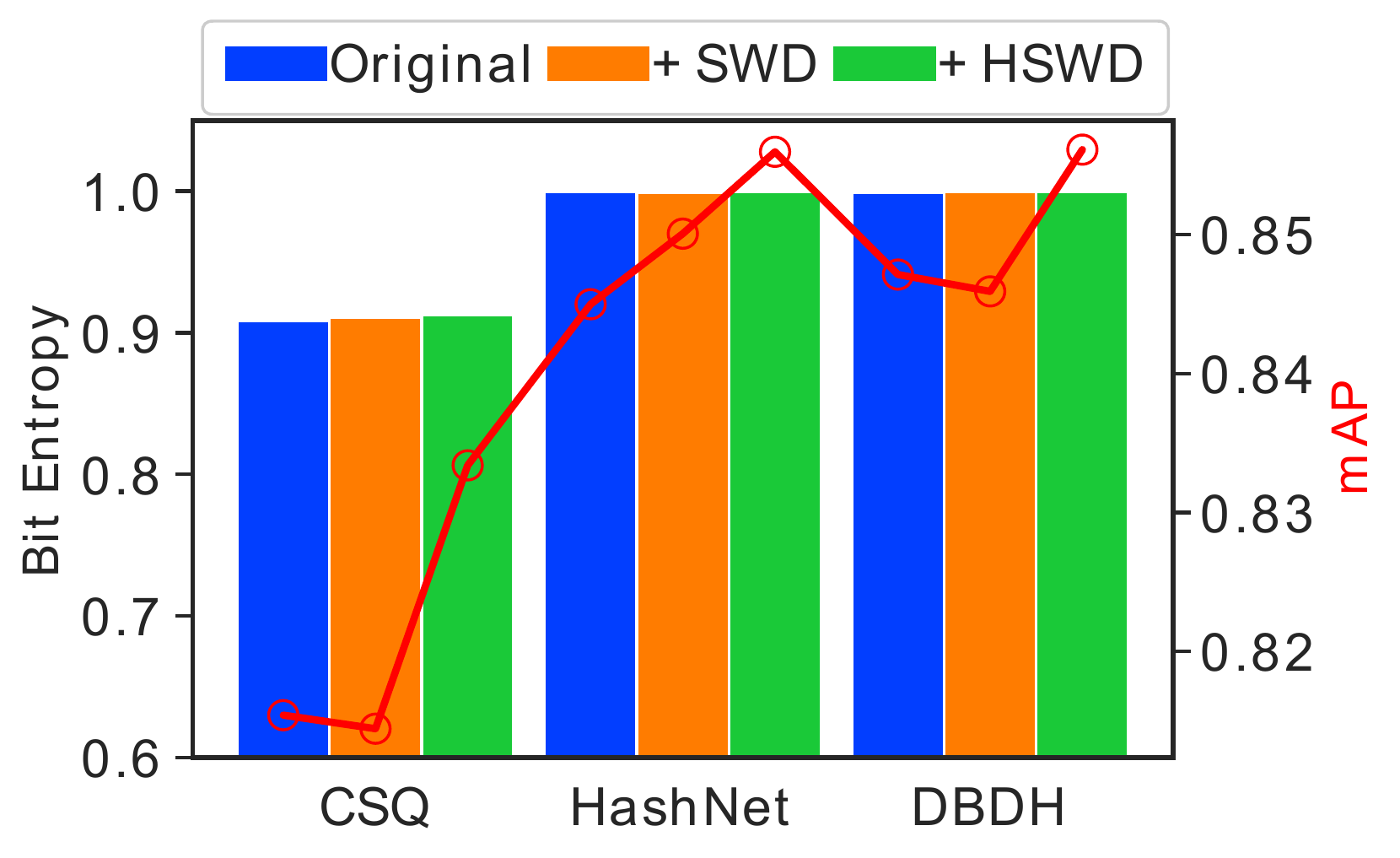}}
}
        \caption{Bit Entropy}\label{fig:1b}
	\end{subfigure}
}
	\caption{Quantization Error (angle, in radian, between continuous codes and corresponding hash codes) and Bit Entropy (CIFAR10).}\label{fig:rebuttal_quantization_constraints}
\end{figure}

The proposed method replaces several quantization losses with a single loss (SWD/HSWD), allowing efficient and effective quantization in deep supervised hashing methods. In Figure~\ref{fig:rebuttal_quantization_constraints}, we show that the proposed quantization approaches simultaneously achieve low-quantization and balanced bits. As we can observe in this figure, when the quantization error is reduced (as in the proposed approaches), the retrieval performance increases. Furthermore, the proposed approaches also consistently achieve highly balanced bits (i.e., high bit entropies). Higher balanced bits also lead to better performance. 

\vspace{0.1in}

\section{Conclusions}\label{sec:conclusions}

In this paper, we proposed a novel single-loss quantization objective for the deep supervised image hashing problem. Different from the existing quantization approaches, our model selects an optimal uniform discrete distribution and directly minimizes the distribution distance between the output distribution of the hash function and this uniform distribution. We considered the Sliced Wasserstein Distance as the choice of the distributional distance for its low-sample complexity and easier estimation. However, the Sliced Wasserstein Distance can require projections into several random directions that do not contain useful information about the separation of the data. By studying the properties of the distributions under consideration, we proposed a variant of SWD, called HSWD, and showed that it only requires a small number of informative directions.  HSWD can achieve significant computational gains than SWD. Our experiments validate that the proposed quantization approach can improve the performance of several representative deep supervised-hashing methods on several datasets. Our work makes one leap towards leveraging an efficient, robust quantization approach for deep supervised hashing and we envision that our model will serve as a motivation for improving other related hashing applications.

\newpage
{\small
\bibliographystyle{ieee_fullname}
\bibliography{egbib,standard}
}

\end{document}